\documentclass{article}
\usepackage{spconf,amsmath,graphicx}

\usepackage{enumitem}
\setlist{nosep, leftmargin=14pt}
\usepackage[most]{tcolorbox}
\newcommand{\eatme}[1]{ }
\usepackage{threeparttable}
\usepackage{listings}
\usepackage{amsmath}
\usepackage{graphicx}
\usepackage{caption}
\usepackage{subcaption}
\usepackage{algorithm}
\usepackage{algpseudocode}
\usepackage{mwe} 

\usepackage{amssymb} 
\usepackage{amsmath} 

\title{\lowercase{h}SNMF: Hybrid Spatially Regularized NMF for Image-Derived Spatial Transcriptomics}
\name{\begin{tabular}{c}
Md Ishtyaq Mahmud$^{\star 1}$ \quad Veena Kochat$^{\star 2}$ \quad 
 Suresh Satpati$^{2}$\thanks{$^\star$ Core contributors, $^{\ddagger}$ Joint advising}   \quad Jagan Mohan Reddy Dwarampudi$^{1}$ \\[3pt]
Humaira Anzum$^{1}$ \quad 
Kunal Rai$^{\ddagger 2}$ \quad Tania Banerjee$^{\ddagger 31}$\thanks{Correspondence to: tbanerjee@uh.edu, krai@mdanderson.org}
\end{tabular}}
\address{$^{1}$ Department of Electrical and Computer Engineering, University of Houston \\
       $^{2}$ Department of Genomic Medicine and MDACC Epigenomics Therapy Initiative (METI), \\
       \hspace*{1em} MD Anderson Cancer Center, Houston TX, USA \\
        $^{3}$ Department of Information Science Technology, University of Houston, Houston, TX, USA}

\begin{document}
%
\maketitle

\begin{abstract}
High-resolution spatial transcriptomics platforms, such as Xenium, generate
single-cell images that capture both molecular and spatial context, but their
extremely high dimensionality poses major challenges for representation
learning and clustering. In this study, we analyze data from the
Xenium platform, which captures high-resolution images of tumor
microarray (TMA) tissues and converts them into cell-by-gene matrices suitable
for computational analysis. We benchmark and extend nonnegative matrix
factorization (NMF) for spatial transcriptomics by introducing two spatially
regularized variants. First, we propose
Spatial NMF (SNMF), a lightweight baseline that enforces local
spatial smoothness by diffusing each cell's NMF factor vector over its spatial
neighborhood. Second, we introduce Hybrid Spatial NMF (hSNMF),
which performs spatially regularized NMF followed by Leiden clustering on a
\emph{hybrid adjacency} that integrates spatial proximity (via a
contact--radius graph) and transcriptomic similarity through a tunable mixing
parameter~$\alpha$. Evaluated on a cholangiocarcinoma dataset, SNMF
and hSNMF achieve markedly improved spatial compactness (CHAOS~$<0.004$,
Moran's~$I>0.96$), greater cluster separability (Silhouette~$>0.12$, DBI~$<1.8$), and higher biological coherence (CMC and enrichment)
compared to other spatial baselines. \textbf{Availability and implementation}: 
https://github.com/ishtyaqmahmud/hSNMF
\end{abstract}

\section{Introduction}

High-resolution spatial transcriptomics (ST) technologies, exemplified by the Xenium platform (10x Genomics), have emerged as transformative tools in genomics \cite{schroyer2024spatial}, \cite{janesick2023high}. By capturing gene expression profiles at single-cell resolution while preserving native spatial coordinates, these assays enable unprecedented analysis of tissue microenvironments, and spatial organization in complex diseases\cite{shang2022spatially, longo2021integrating}. However, the resulting data are extremely high-dimensional and sparse, posing major computational challenges \cite{fang2023computational}. As in single-cell RNA sequencing (scRNA-seq), effective dimensionality reduction (DR) is critical to reveal cellular subpopulations\cite{du2023advances}. Because ST integrates quantitative imaging with high-throughput genomics, advances in this area demand the same methodological rigor that has driven progress in biomedical image analysis: precise signal extraction, robust dimensionality reduction, and interpretable feature representations. Consequently, enhancing spatial interpretability and scalability enables data-driven insights into tissue structure and disease pathology at single-cell resolution \cite{zhong2024interpretable, kleino2022computational}.

Standard DR methods such as Principal Component Analysis (PCA) and Non-negative Matrix Factorization (NMF) \cite{lee1999learning} treat each cell as an independent sample, ignoring spatial context. This ``a-spatial'' assumption often produces fragmented, ``salt-and-pepper'' clusterings that are biologically incoherent and fail to capture continuous tissue structures. To address this limitation, spatially aware DR methods have been developed. Broadly, these approaches fall into two major families:
\begin{enumerate}
    \item \textbf{Probabilistic Models}, such as Nonnegative Spatial Factorization (NSF) \cite{townes2021nonnegative}, which use Gaussian Processes to model spatial correlation. While powerful, these models are computationally intensive and scale poorly with the large cell counts generated by modern platforms like Xenium.
    \item \textbf{Filter-based Methods}, including Randomized Spatial PCA (RASP) \cite{gingerich2025randomized} and our baseline Spatial NMF (SNMF), which first learn an a-spatial embedding (e.g., via NMF) and then smooth it using a spatial graph. These are efficient and scalable but often rely on oversimplified k-nearest neighbor (k-NN) topologies that fail to reflect true physical proximity or multi-scale tissue structure.
\end{enumerate}

Deep learning–based approaches such as graph neural networks (GNNs) have also been proposed, but their latent embeddings are often opaque and less interpretable for biological analysis \cite{sun2024comprehensive}, \cite{yang2024graphpca}.

This work introduces two spatially aware extensions of NMF for dimensionality reduction in spatial transcriptomics. The key contributions of the paper may be summarized as:
\begin{enumerate}
    \item We present Spatial NMF (SNMF), a lightweight two-stage baseline that enhances standard NMF embeddings through spatial smoothing of cell factors over local neighborhoods, providing an interpretable bridge between expression-based and spatially informed analyses.
    \item We further propose Hybrid Spatial NMF (hSNMF), which performs spatially regularized NMF followed by Leiden clustering on a \emph{hybrid adjacency} that combines spatial proximity and transcriptomic similarity. The spatial component employs a contact--radius graph that merges short-range contact edges with longer-range contextual connections, allowing hSNMF to jointly optimize geometric contiguity and molecular coherence.
\end{enumerate}
Together, SNMF and hSNMF offer parameter-efficient, interpretable, and scalable
baselines for capturing spatial organization in high-dimensional gene
expression data.

\section{Methodology} 
\subsection{Dataset and Preprocessing\label{sec:pp}}
We used the same Xenium spatial transcriptomics (ST) dataset and initial
preprocessing pipeline described in our prior work~\cite{mahmud2025benchmarking},
with additional steps to construct spatial graphs for the proposed SNMF and
hSNMF models. Briefly, the dataset consists of \textit{N~=~25} cholangiocarcinoma
patients (total of \textit{M~=~40} tumor microarray cores) profiled on the
Xenium platform using a 480-gene target panel, yielding $\approx$212{,}000
cells. Each core produced high-resolution tissue images that were processed
to generate single-cell expression matrices and spatial coordinates. All
samples were anonymized TMAs obtained from the MD Anderson Cancer Center,
representing intrahepatic cholangiocarcinoma resections.

\begin{enumerate}
    \item \textit{Quality control (QC) and gene filtering.}
    Genes detected in fewer than three cells and cells with fewer than 200
    detected genes (including negative-control probes) were removed. Doublet
    detection was performed with \texttt{Scrublet} following Wolock
    et~al.~\cite{WOLOCK2019281}, and cells with doublet scores $>$0.2 were
    excluded. After QC, 191{,}125 high-confidence single-cell profiles remained.

    \item \textit{Normalization and transformation.}
    Counts were normalized per cell to 10{,}000 total counts and log-transformed
    as $\log_e(x{+}1)$.

    \item \textit{Spatial encoding.}
    Each cell’s centroid $(x,y)$ was extracted from the Xenium metadata and used
    to build spatial graphs. Specifically, we constructed both a
    short-range \textbf{contact graph} (radius~=~20~$\mu$m) and a broader
    \textbf{radius graph} (radius~=~80~$\mu$m), later combined into a
    hybrid adjacency for spatial smoothing and clustering.

\end{enumerate}

\subsection{Existing Spatial Dimentionality Reduction Methods}
To evaluate the performance of our proposed hSNMF framework, we conducted a comparative analysis against several state-of-the-art methods. These methods utilize different underlying technologies, including PCA and matrix factorization.
\paragraph* {RASP (Randomized Spatial PCA)~\cite{gingerich2025randomized}:}

RASP is a spatially aware dimensionality reduction method optimized for large-scale ST data. Built on a randomized two-stage PCA framework with sparse matrix operations, it achieves high computational efficiency and scales to hundreds of thousands of spatial locations. A configurable spatial smoothing step integrates spatial context, enabling de-noised expression reconstruction and flexible inclusion of non-transcriptomic covariates.    
\paragraph*{NSF (Nonnegative Spatial Factorization)~\cite{townes2021nonnegative}:} NSF is a spatially aware probabilistic model that extends NMF by placing Gaussian Process priors on latent factors, enforcing spatial smoothness. Gene counts are modeled with non-negative loadings under Poisson or Negative Binomial likelihoods, producing sparse, interpretable components. A hybrid variant (NSFH) combines spatial and nonspatial factors to quantify gene-level spatial effects.
\subsection{Proposed Spatial Extensions of NMF}
\paragraph*{Spatial NMF (SNMF).}
We introduce \emph{Spatial NMF (SNMF)} as a lightweight two-stage baseline that incorporates spatial context into standard nonnegative matrix factorization (NMF).

Given the log-normalized cell-by-gene expression matrix
$\mathbf{X} \in \mathbb{R}_{\ge 0}^{n \times p}$, we first compute nonnegative latent factors
$\mathbf{W} \in \mathbb{R}_{\ge 0}^{n \times k}$ and
$\mathbf{H} \in \mathbb{R}_{\ge 0}^{k \times p}$ by solving
\begin{equation}
\min_{\mathbf{W}, \mathbf{H} \ge 0}
\; \lVert \mathbf{X} - \mathbf{W}\mathbf{H} \rVert_F^2.
\end{equation}

To encourage spatial smoothness, we apply a single-step post hoc averaging to the latent representation.
Specifically, each cell’s factor vector $\mathbf{w}_i$ is replaced by the average of its $k=15$ nearest spatial neighbors:
\begin{equation}
\mathbf{w}'_i =
\frac{1}{|\mathcal{N}_S(i)|}
\sum_{j \in \mathcal{N}_S(i)} \mathbf{w}_j,
\end{equation}
where $\mathcal{N}_S(i)$ denotes the spatial neighborhood of cell $i$.
This local averaging acts as a low-pass filter on the latent space, yielding a smoothed embedding
$\mathbf{W}'$ that is used for downstream clustering.

SNMF is included as a conceptual baseline to illustrate the effect of simple spatial smoothing on NMF embeddings; however, we focus our empirical evaluation on hSNMF, which consistently outperformed SNMF in preliminary analyses.

\begin{table*}[ht]
\footnotesize
\centering
\caption{Pareto-optimal hyperparameter settings per method and their performance metrics.}
\vspace{-2mm}
\label{tab:best_configs}
\begin{tabular}{|l|p{0.7cm}|p{0.6cm}|p{1cm}|p{1.3cm}|p{1.5cm}|p{1.7cm}|p{0.8cm}|p{1cm}|p{1cm}|p{1.2cm}|}
\hline
Method 
  & $k$
  & $\rho$
  & Clusters
  & CHAOS $\downarrow$
  & Moran's I $\uparrow$
  & Silhouette $\uparrow$
  & DBI $\downarrow$
  & CMC $\uparrow$
  & MER $\downarrow$
  & Enrich. $\uparrow$ \\
\hline
\hline
RASP  
  & 5 
  & 0.4  
  & 17
  & 0.005 
  & 0.649 
  & 0.166  
  & 1.399  
  & 0.796
  & 0.548
  & 1.968 \\

\hline
RASP  
  & 10 
  & 0.4  
  & 15
  & 0.004
  & 0.649
  & 0.183  
  & 1.604 
  & 0.838
  & 0.385 
  & 2.078 \\
\hline
RASP  
  & 15 
  & 0.4  
  & 14
  & 0.004 
  & 0.649 
  & 0.166  
  & 1.711 
  & 0.809
  & 0.276
  & 1.729 \\

\hline
\hline
NSF 
  & 5 
  & 0.4  
  & 24 
  & 0.003 
  & 0.992 
  & -0.879
  & 2.043
  & 0.580  
  & 0.934
  & 1.475 \\
  
\hline
NSF 
  & 5 
  & 0.6  
  & 29
  & 0.003
  & 0.992
  & -0.881
  & 3.191
  & 0.587  
  & 1.475
  & 0.927 \\
\hline
NSF 
  & 10 
  & 0.6  
  & 25 
  & 0.005
  & 0.993
  & -0.585
  & 5.857
  & 0.521  
  & 1.764
  & 0.954 \\
\hline
\hline
SNMF 
  & 10  
  & 0.4  
  & 27
  & 0.003  
  & 0.964
  & 0.128
  & 1.207  
  & 0.764
  & 1.876  
  & 0.555 \\
\hline
SNMF 
  & 10  
  & 0.8  
  & 39
  & 0.003  
  & 0.964 
  & 0.153 
  & 1.335  
  & 0.762
  & 1.957  
  & 0.586 \\
\hline
SNMF
  & 10 
  & 1  
  & 43
  & 0.003  
  & 0.964 
  & 0.163 
  & 1.294  
  & 0.760
  & 1.843  
  & 0.599 \\
\hline
\hline
hSNMF
  & 10  
  & 0.4  
  & 23
  & 0.002
  & 0.970
  & 0.274
  & 1.423  
  & 0.720
  & 0.445
  & 1.967 \\
  \hline
hSNMF
  & 15  
  & 0.4  
  & 24
  & 0.002
  & 0.982
  & 0.178
  & 1.689  
  & 0.715
  & 0.505
  & 2.285 \\
  \hline
  hSNMF
  & 25  
  & 0.4  
  & 27
  & 0.002
  & 0.941
  & 0.228
  & 1.687  
  & 0.721
  & 0.497
  & 2.101 \\
  \hline
  hSNMF
  & 25  
  & 0.8  
  & 29
  & 0.002
  & 0.941
  & 0.229
  & 1.653  
  & 0.717
  & 0.981
  & 2.175 \\
  \hline
\end{tabular}
\vspace{-3mm}
\end{table*}

\paragraph*{Hybrid Spatially Smoothed NMF (hSNMF).}
\label{subsec:hsnmf}
We next propose \emph{Hybrid Spatially Smoothed NMF (hSNMF)}, which extends SNMF by applying iterative spatial diffusion over a hybrid spatial graph and integrating spatial and molecular similarity during clustering.

We begin with the same log-normalized expression matrix
$\mathbf{X} \in \mathbb{R}_{\ge 0}^{n \times p}$ and factorize it using standard NMF with $k$ components
(\textit{init}=\texttt{nndsvda}, \textit{max\_iter}=500, \textit{random\_state}=0):
\begin{equation}
\mathbf{X} \approx \mathbf{W}\mathbf{H}.
\end{equation}

To incorporate spatial coherence, we construct a \emph{hybrid spatial adjacency matrix} $\mathbf{A}_s$ that combines
contact-based edges within radius $r_c = 20\,\mu\mathrm{m}$ and
radius-based edges within $r_r = 80\,\mu\mathrm{m}$,
merged via maximum-weight selection to preserve local tissue structure.
Let $\mathbf{D}$ denote the degree matrix of $\mathbf{A}_s + \mathbf{I}$.
We define a row-stochastic diffusion operator
\begin{equation}
\mathbf{P} = \mathbf{D}^{-1}(\mathbf{A}_s + \mathbf{I}).
\end{equation}

Spatial smoothing is then applied to the latent factors through iterative diffusion:
\begin{equation}
\mathbf{W}^{(t+1)} =
(1 - \beta)\,\mathbf{W}^{(t)} +
\beta\,\mathbf{P}\mathbf{W}^{(t)},
\end{equation}
with $\beta = 0.8$ and two diffusion steps, producing the spatially smoothed embedding $\mathbf{W}_s$.
To preserve nonnegativity, we clip any small numerical negatives in $\mathbf{W}_s$ to zero.

Next, we construct a $k_{\mathrm{nn}}$-nearest-neighbor graph $\mathbf{A}_f$ in the smoothed latent space $\mathbf{W}_s$ to capture transcriptional similarity ($k_{\mathrm{nn}}=15$).
We then form a \emph{dual-graph adjacency}
\begin{equation}
\mathbf{A}_{\text{mix}} =
\alpha\,\mathbf{A}_s + (1 - \alpha)\,\mathbf{A}_f,
\end{equation}
where $\alpha \in [0,1]$ controls the trade-off between spatial proximity and molecular similarity ($\alpha = 0.5$ by default).
All adjacency matrices are row-stochastic to ensure scale comparability.

Finally, Leiden clustering is applied directly to $\mathbf{A}_{\text{mix}}$ to obtain communities that are both spatially contiguous and transcriptionally coherent.
The resulting clusters are evaluated using spatial (CHAOS, Moran’s $I$), geometric (Silhouette, DBI), and biological (Marker Fraction, MER, Enrichment) metrics.

\section{Evaluation Metrics}\label{sec:metrics}
We quantitatively assessed the spatial quality of embeddings using two complementary metrics: CHAOS, which measures spatial compactness of discrete clusters, and Moran’s I, which measures spatial autocorrelation of continuous latent factors.

\paragraph*{CHAOS: Spatial Cluster Compactness:}
CHAOS quantifies how spatially cohesive each cluster is by computing the mean distance between each cell and its nearest same-cluster neighbor. Lower values indicate tighter, more spatially compact clusters. Given standardized $(x, y)$ coordinates for all $N$ cells, the CHAOS score is:
\[
\text{CHAOS} = \frac{1}{N} \sum_{k \in K} \sum_{i \in C_k} 
\min_{j \in C_k, j \neq i} d(i, j),
\]
where $K$ is the set of clusters and $C_k$ the set of cells in cluster $k$.

\begin{figure*}[t]
  \centering
  \begin{subfigure}[t]{0.3\textwidth}
    \centering
    \includegraphics[clip, trim=8cm 0cm 5cm 0cm, width=\columnwidth]{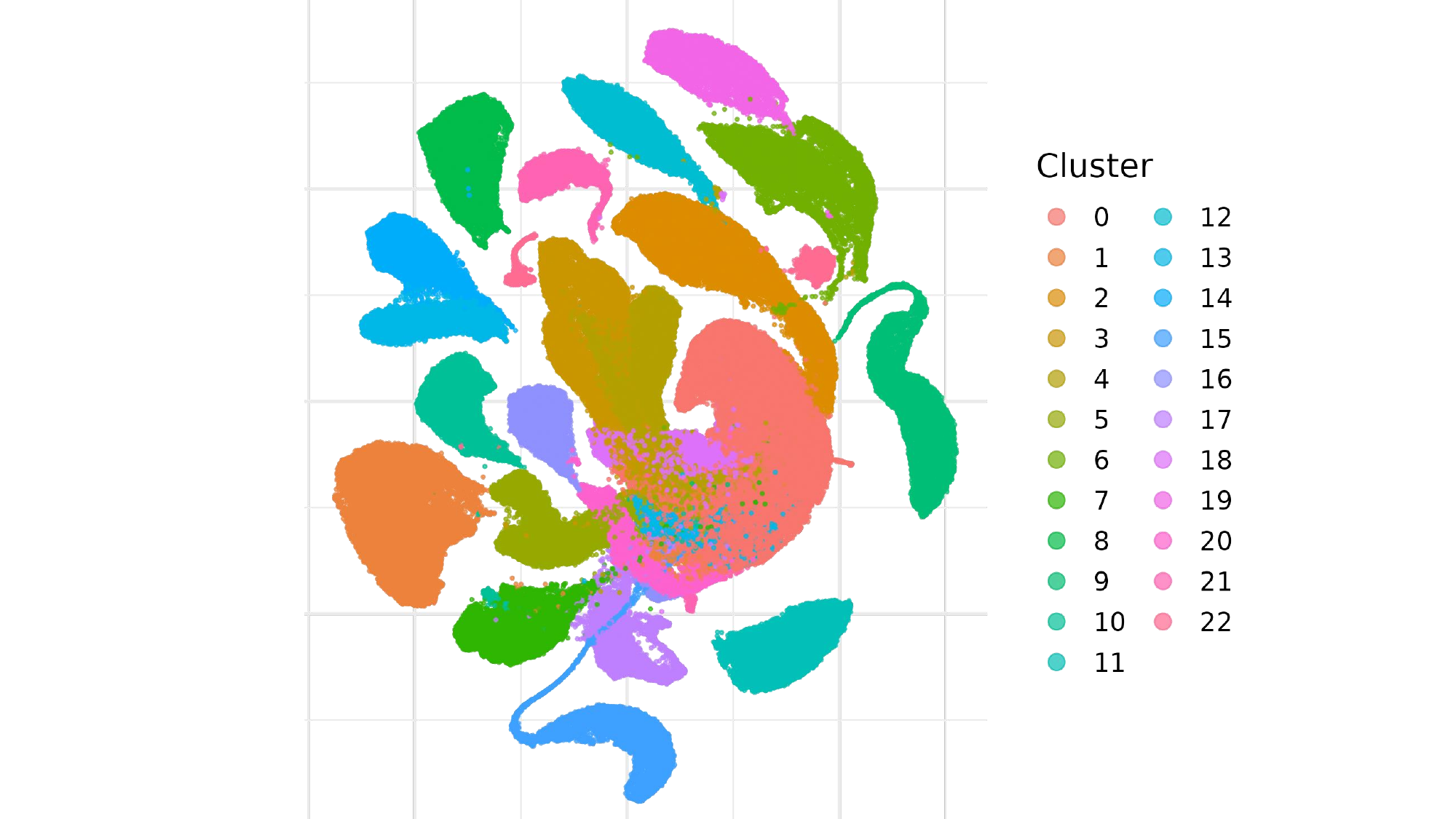}
    \caption{hSNMF UMAP projection}
    \label{fig:pareto_hsnmf}
  \end{subfigure}
  \begin{subfigure}[t]{0.3\textwidth}
    \centering
    \includegraphics[clip, trim=10cm 0cm 5cm 0cm,width=\linewidth]{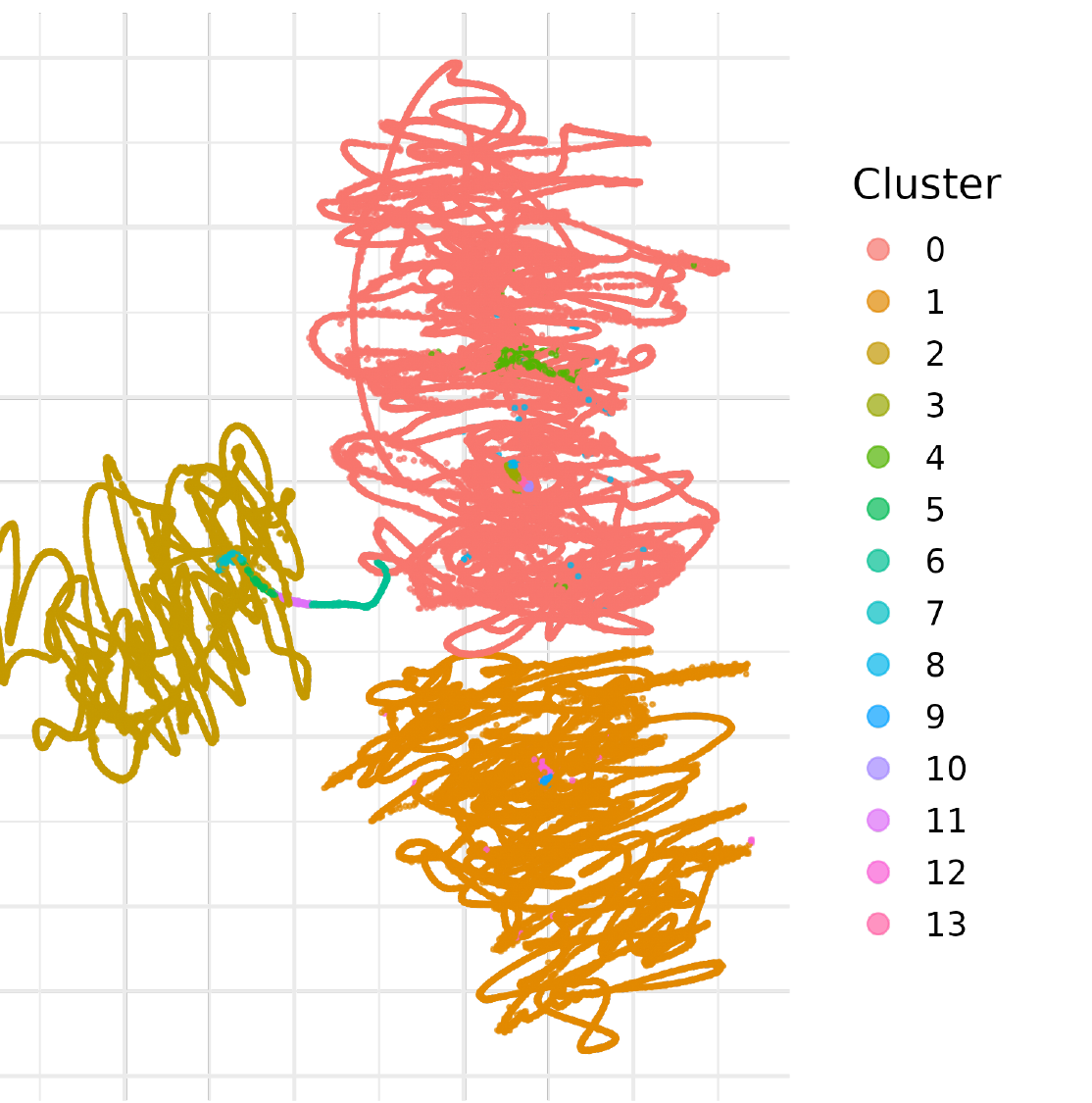}
    \caption{NSF UMAP projection}
    \label{fig:pareto_nsf}
  \end{subfigure}
  \begin{subfigure}[t]{0.3\textwidth}
    \centering
    \includegraphics[clip, trim=6.6cm 0cm 3cm 0cm, width=\columnwidth]{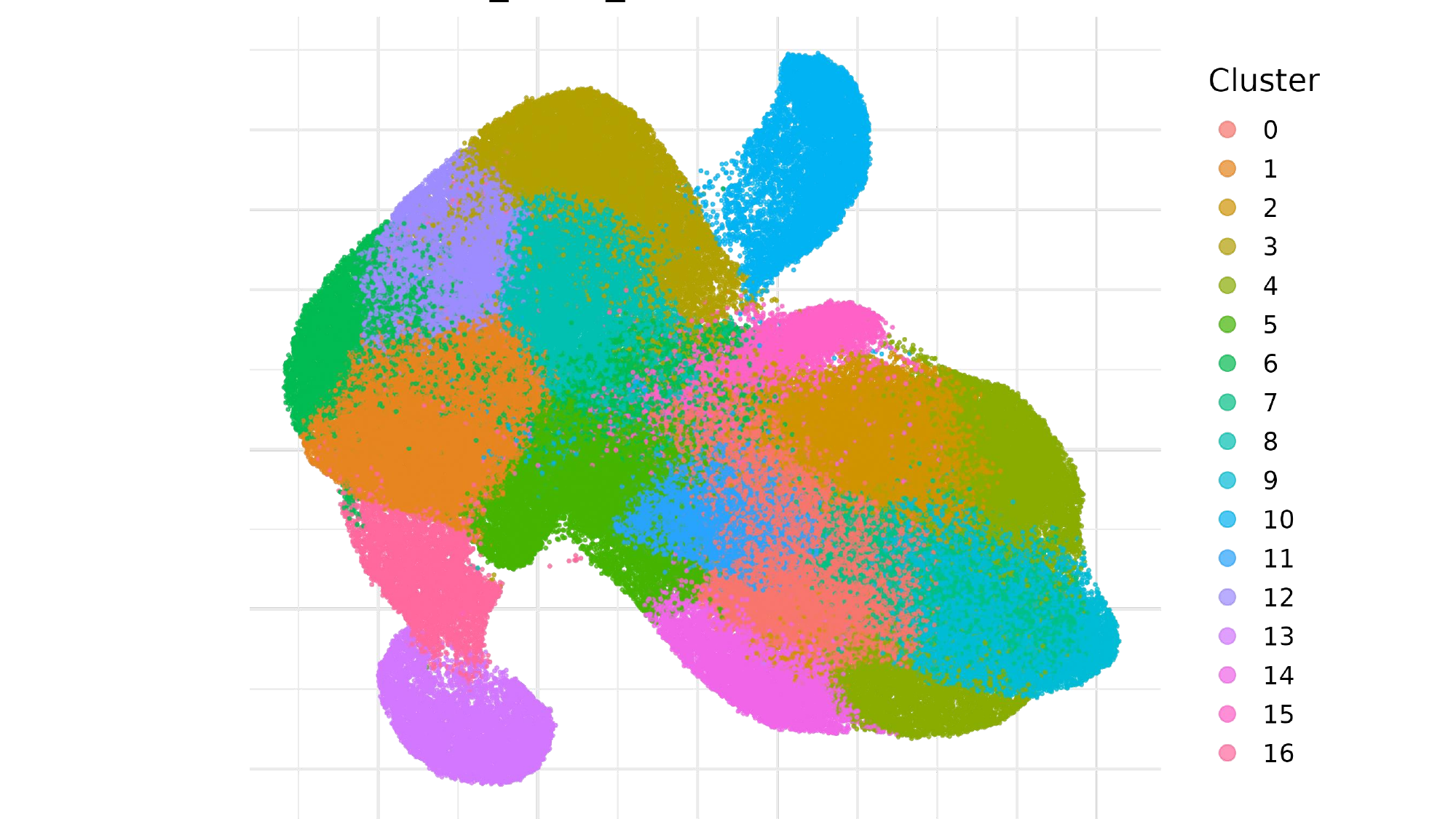}
    \caption{RASP UMAP projection}
    \label{fig:pareto_rasp}
  \end{subfigure}

  \vspace{-2mm}

  \caption{UMAP projections of cell clusters obtained after applying Leiden clustering on the reduces cellxgene vectors.}
  \label{fig:umap}
  \vspace{-5mm}
\end{figure*}
\paragraph*{Moran’s I: Spatial Autocorrelation:}
Moran’s I measures the spatial correlation of continuous features (e.g., latent factors). For feature vector $y$ and spatial weight matrix $W$, it is defined as:
\[
I = \frac{N}{\sum_i \sum_j w_{ij}}
\frac{\sum_i \sum_j w_{ij}(y_i - \bar{y})(y_j - \bar{y})}
{\sum_i (y_i - \bar{y})^2},
\]
where $N$ is the number of cells, $y_i$ the feature value for cell $i$, and $w_{ij}$ the spatial weight (from a $k$-NN or radius graph). Values $I>0$, $I<0$, and $I\approx0$ correspond to clustered, dispersed, and random spatial patterns, respectively. We compute $I$ on the first latent factor of each embedding for cross-model comparison.

\paragraph*{Other Metrics}
In addition to CHAOS and Moran's I, we also evaluate embeddings using several metrics defined by Mamud et. al in~\cite{mahmud2025benchmarking}: Silhouette Score, Davies–Bouldin Index (DBI), Cluster Marker Coherence (CMC), Marker Exclusion Rate (MER), and Marker Enrichment.

\section{Results} 
\subsection{Pareto-front Analysis}
To systematically evaluate the performance trade-offs inherent in each spatial dimensionality reduction method, we performed a comprehensive grid search across key hyperparameters. For each method (RASP, NSF, SNMF, and hSNMF), we varied the latent dimensionality
\[
k \in \{5, 10, 15, 20, 25, 30, 35, 40\}
\]
and Leiden clustering resolution 
\[
\rho \in \{0.1, 0.2, 0.3, 0.4, 0.5, 0.6, 0.7, 0.8, 0.9, 1.0, 1.2\}.
\]

At each $(k, \rho)$ parameter combination, we evaluated the resulting embedding and clustering using the suite of quantitative metrics defined in Section~\ref{sec:metrics}. We then identified the Pareto-optimal $(k, \rho)$ configurations for each method. The Pareto front, in our case, represents the set of hyperparameter settings for which no other setting simultaneously improves CHAOS, Moran's I, Silhouette Scores, and DBI.

Table~\ref{tab:best_configs} summarizes the Pareto-optimal configurations and
corresponding performance metrics for all evaluated spatial DR variants. The proposed spatial models, \textbf{SNMF} and
\textbf{hSNMF}, consistently exhibit high spatial compactness
(CHAOS~$<0.004$) and strong spatial autocorrelation
(Moran's~$I \approx 0.96$--$0.98$), confirming effective spatial
regularization. Unlike the NSF baseline, which yields
negative Silhouette scores and poor cluster separability, SNMF and hSNMF
achieve positive Silhouette values (0.15--0.27) with moderate
Davies--Bouldin indices, indicating well-separated and cohesive clusters. hSNMF, which incorporates the contact--radius hybrid graph, further
improves both Silhouette and enrichment while maintaining low MER and high
CMC, thereby balancing spatial contiguity and molecular coherence.
In contrast, the RASP baseline attains modest geometric performance
but markedly lower Moran's~$I$, reflecting weaker spatial consistency.
Overall, the hybrid hSNMF formulation attains the most favorable trade-off
among spatial smoothness, cluster compactness, and biological fidelity.

Figure~\ref{fig:umap} compares the UMAP projections of cell embeddings 
obtained from hSNMF, NSF, and RASP. The proposed hSNMF method 
(Fig.~\ref{fig:umap}a) produces compact, well-separated clusters with smooth 
boundaries, indicating both geometric cohesion and clear transcriptional 
differentiation. In contrast, the non-spatial NSF baseline 
(Fig.~\ref{fig:umap}b) yields elongated, overlapping clusters lacking spatial 
organization, consistent with its low Silhouette scores and negative 
Moran’s~$I$. The RASP embedding (Fig.~\ref{fig:umap}c) preserves 
some global structure but exhibits partial mixing between clusters. Overall, 
hSNMF yields the most coherent and separable manifolds, aligning with its 
superior quantitative performance in Table~\ref{tab:best_configs}.
\section{Conclusion}
We presented two spatially regularized extensions of nonnegative matrix
factorization (NMF) for dimensionality reduction in spatial transcriptomics.
The proposed Spatial NMF (SNMF) introduces spatial smoothness through
local graph diffusion of cell factors, providing a simple yet effective way to
enhance tissue coherence in low-dimensional embeddings. Building on this, the
Hybrid Spatial NMF (hSNMF) (implemented using a contact--radius
hybrid graph) integrates both spatial proximity and transcriptomic similarity
within a unified clustering framework. Experiments on a Xenium
cholangiocarcinoma dataset demonstrate that SNMF and hSNMF substantially
improve spatial compactness (low CHAOS, high Moran's~$I$) and biological
consistency (higher CMC and enrichment) compared to non-spatial baselines.
The hybrid formulation achieves the most balanced trade-off between geometric
and molecular coherence, yielding spatially contiguous yet biologically
interpretable and separable clusters. Future work will extend this framework to incorporate
multi-scale spatial graphs and compare against deep generative spatial models.

\vspace{-3mm}
\section{Compliance with Ethical Standards}
\vspace{-1mm}
This retrospective study utilized de-identified Xenium-based spatial transcriptomics data from patients evaluated for cholangiocarcinoma at the MD Anderson Cancer Center (MDACC). The study protocol was approved by the MDACC Institutional Review Board (IRB). 
Given the retrospective nature and use of de-identified data, the requirement for informed consent was waived per IRB determination.

\vspace{-3mm}
\section{Acknowledgement}
\vspace{-1mm}
The work was, in part, supported by STRIDE funding to METI from MD Anderson Cancer Center. We thank the cholangiocarcinoma clinical and pathology group at MD Anderson Cancer Center for providing the tissue microarrays. This project was also supported by the Center for Transformative Pathology and Health (CTPH) under award UM1TR004539. The authors declare that they have no relevant financial or non-financial conflicts of interest.

\bibliographystyle{IEEEbib}
\bibliography{strings,ISBI_refs}

\end{document}